\documentclass[letterpaper,10pt,conference]{ieeeconf}
\IEEEoverridecommandlockouts
\usepackage{times}
\usepackage{cite}
\usepackage[dvipsnames]{xcolor}
\usepackage{multicol}
\usepackage{multirow}
\usepackage{pifont}
\usepackage{amssymb}
\usepackage{amsmath}
\usepackage{import}
\usepackage{graphicx}
\usepackage{booktabs}
\usepackage{makecell}
\usepackage{tabularx}
\usepackage{bm}
\usepackage[normalem]{ulem}
\useunder{\uline}{\ul}{}
\usepackage{algorithm}
\usepackage{algpseudocode}
\algtext*{EndFor}\algtext*{EndWhile}\algtext*{EndIf}
\usepackage[hidelinks]{hyperref}
\usepackage{eso-pic}
\newcommand{\ieeenotice}{\AddToShipoutPictureBG*{\AtPageUpperLeft{\put(0,-30){\makebox[\paperwidth]{\footnotesize This work has been submitted to the IEEE for possible publication. Copyright may be transferred without notice, after which this version may no longer be accessible.}}}}}

\newcommand{\method}{WholeBodyMem}
\title{Memory before Goal:\\ Role of Pre-History Memory for Humanoid Loco-Manipulation}
\author{Chenyuan Chen, Ran Ding, Lechen Zhang, Zewen He and Yoshihiko Nakamura\\
Mohamed bin Zayed University of Artificial Intelligence (MBZUAI), Abu Dhabi, UAE\\
{\small\texttt{\{chenyuan.chen, yoshihiko.nakamura\}@mbzuai.ac.ae}}\vspace{-1.5em}
}

\begin{document}
\ieeenotice
\maketitle
\thispagestyle{empty}
\pagestyle{empty}

\begin{abstract}
Humanoid loco-manipulation in dynamic scenes is non-Markovian: cues vanish as objects change or are hidden, or leave the view as the robot turns and walks, yet whole-body vision-language-action (VLA) policies see only the present. We present \method{}, an object-guided perceptual memory that keeps the pre-history of object events seen before the robot acts on its goal. SAM3 tracks the task objects, and their events causally select key frames within an adaptive budget, which the action expert reads by cross-attention together with object tokens encoding appearance, geometry and time. We also contribute HuMeGo, a humanoid memory benchmark with six simulated and six real-world Unitree G1 tasks. Keeping the right frames rather than more frames, \method{} surpasses memoryless and uniform-memory baselines, raising its base policy's success from 29.1\% to 72.6\% in simulation and from 36.7\% to 88.3\% on the real robot, while also improving the standard SIMPLE tasks (70.2\% to 86.9\%). Project~page:~\mbox{\href{https://memory-before-goal.github.io}{https://memory-before-goal.github.io}}\\
GitHub:~\mbox{\href{https://github.com/crystalccy1/MemoryBeforeGoal}{crystalccy1/MemoryBeforeGoal}}
\end{abstract}

\vspace{-10pt}

\section{Introduction}
\label{sec:intro}

\begin{figure}[t]
  \centering
  \includegraphics[width=\columnwidth]{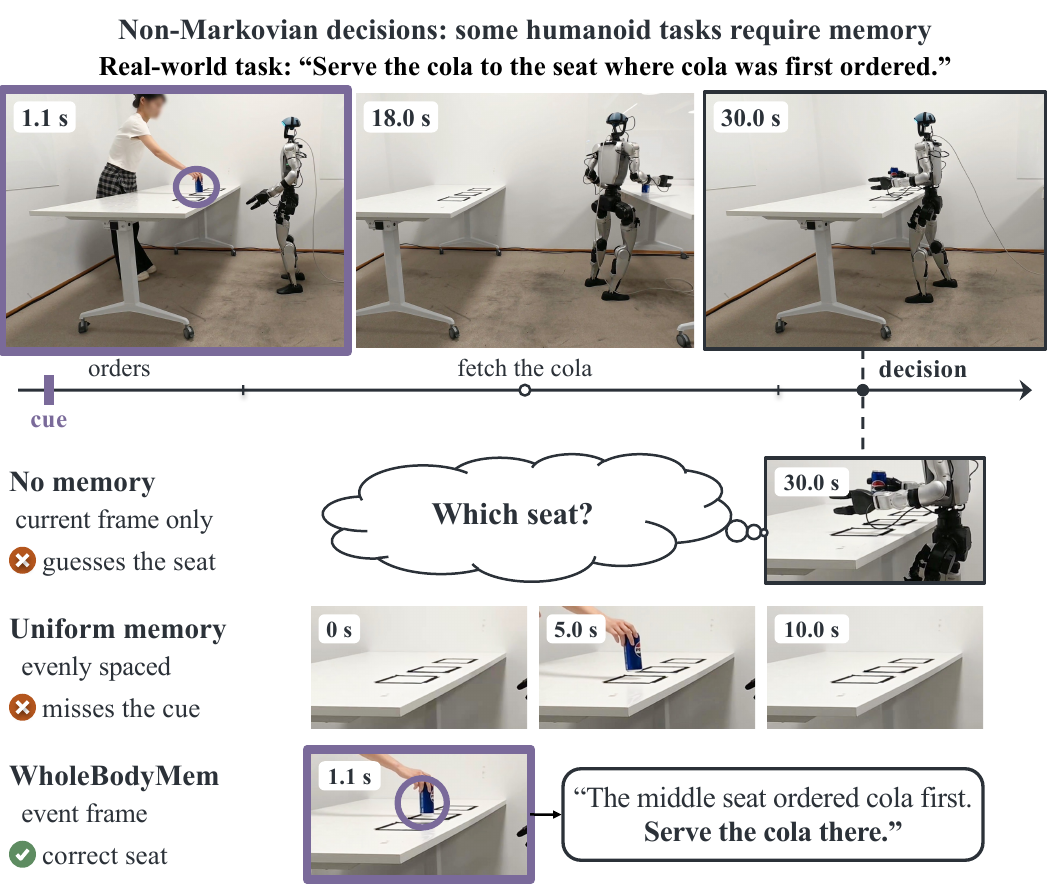}
  \caption{Memory in a real whole-body task. Top: the cola stops first on the middle seat (1.1\,s), then a sprite and the cola stop on the other two; the robot turns away to fetch the cola and, when it returns at 30.0\,s, the three seats look the same.}
  \label{fig:teaser}
  \vspace{7pt}
\end{figure}

Whole-body vision-language-action (VLA) models~\cite{wei2026psi0,bjorck2025gr00t,jiang2026wholebodyvla}, paired with learned whole-body controllers~\cite{li2025amo,luo2025sonic}, have advanced humanoid loco-manipulation considerably. Yet most of them act on the current observation alone, ignoring temporal dependencies, and thus fail in dynamic scenes where objects change state or are hidden~\cite{vanhoorick2023tracking,xia2026targo}, moved or removed. Everyday cues are brief and seldom repeated: a guest waves from one of several identical tables while a waiter carries a tray, and when the waiter turns back, the hand is already down. Fig.~\ref{fig:teaser} shows this situation on a real G1: a cola stops on two seats and a sprite on one, the robot turns away to fetch the cola, and when it comes back the seats look the same again, so a policy that sees only the current view cannot tell where the cola stopped first. Loco-manipulation in dynamic scenes is thus non-Markovian: earlier events determine later decisions, which calls for memory. We refer to it as \emph{pre-history memory}: the record of the object events seen before the robot acts on its goal. Such memory is especially hard for a humanoid, whose head camera moves as it turns and walks, so cues leave the view, and whose memory must be updated causally while whole-body control runs. Even dual-system designs~\cite{bjorck2025gr00t}, in which a VLM planner acts as System~2, can reason over the history only if they locate the key frames. Existing memory strategies, mostly developed for tabletop arms, retain long frame histories~\cite{torne2026mem,jang2025contextvla}, which are costly and dilute brief cues; compress them~\cite{shi2026memoryvla,wang2026nativemem}, losing detail; sample frames at fixed intervals~\cite{dai2026robomme}, missing brief cues; or select event frames tied to the robot's own operations or learned from demonstrations~\cite{zeng2026kemo,yang2026eventvla}.

\begin{figure*}[t]
  \vspace*{5pt}
  \centering
  \includegraphics[width=\textwidth]{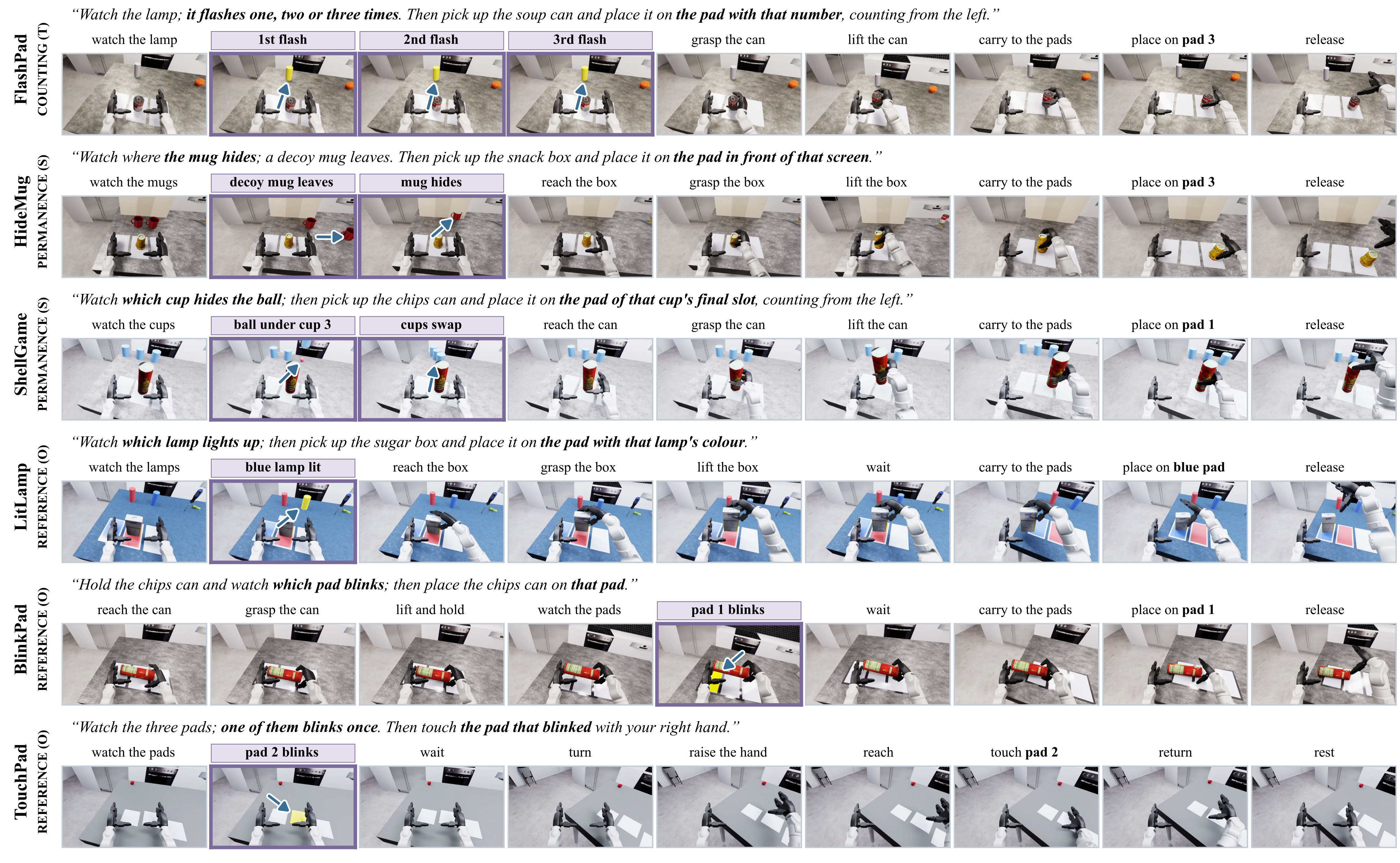}
  \caption{Memory-dependent tasks in simulation, grouped by memory type. Each block shows one task and its memory suite (as in RoboMME~\cite{dai2026robomme}; Sec.~\ref{sec:bench_setting}): the instruction, nine subtasks and the robot's head-camera view at each of them. Bold text marks the cue and the answer; purple borders mark the cue frames, and a blue arrow points at the cue object.}
  \label{fig:tasks}
\end{figure*}

As humans remember the moments when the things they care about change rather than every moment~\cite{zacks2007event}, we propose \method{}, an object-guided perceptual memory for whole-body VLAs. First, a frozen SAM3~\cite{carion2026sam}, a promptable video segmentation model, tracks the task objects in the head-camera stream, and we detect when they appear, disappear, move or change shape, compensating for the camera's own motion. Second, these object events causally select the key frames that fill an adaptive memory budget, stored as native backbone tokens along with mask-pooled object tokens. Finally, the action expert reads this memory via cross-attention and FiLM, so that whole-body actions depend on what happened, not only on what is visible.

Existing humanoid benchmarks~\cite{sferrazza2024humanoidbench,wei2026simple,wang2026humanoidarena,lin2026humanoidmimicgen} keep the goal in view and do not test this problem, whereas memory benchmarks~\cite{dai2026robomme,lei2026robomemarena} are tabletop manipulation. We therefore introduce a memory-dependent humanoid loco-manipulation benchmark. Our contributions are as follows:

\begin{itemize}
  \item We propose \method{}, an object-guided perceptual memory for whole-body VLAs, in which object events, rather than even sampling, select the key frames of an adaptive budget, read together with object tokens.
  \item We introduce HuMeGo, a memory-dependent humanoid loco-manipulation benchmark: six simulated tasks covering object, spatial and temporal memory, where the scene ends the cue and decisions are scored apart from manipulation, and six real-world G1 tasks, where the robot also turns away from the cue.
  \item \method{} keeps the right frames rather than more frames, outperforms memoryless and uniform-memory baselines in decisions and success, helps even on SIMPLE, and carries over to a real G1.
\end{itemize}

\section{Related Work}
\begin{figure*}[t]
  \centering
  \includegraphics[width=\textwidth]{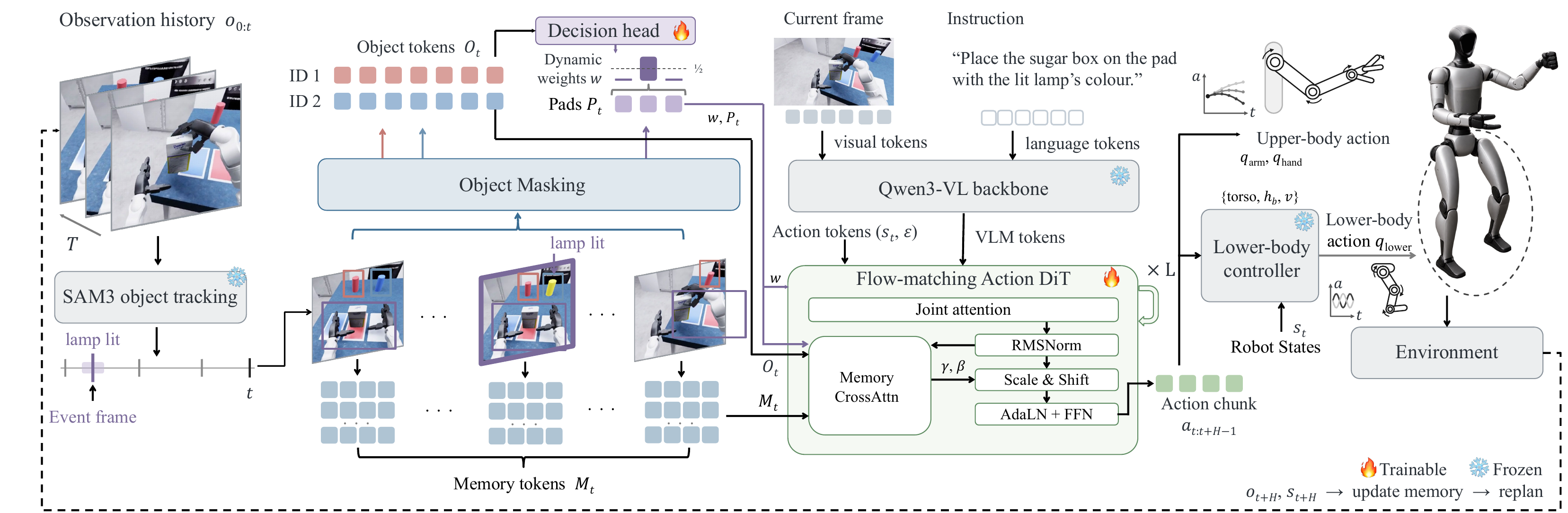}
  \caption{Overview of \method{}. $\times L$: the $L{=}6$ DiT blocks of the action expert.}
  \label{fig:pipeline}
\end{figure*}
\label{sec:related}

\subsection{Whole-Body Humanoid Loco-Manipulation VLAs}
\label{sec:related_wbc}
Whole-body humanoid control uses learned whole-body controllers such as HOVER~\cite{he2025hover}, AMO~\cite{li2025amo} and SONIC~\cite{luo2025sonic}, and demonstrations are collected by whole-body teleoperation~\cite{ze2025twist}. On top of such controllers, vision-language-action (VLA) policies map images and instructions to whole-body actions. Generalist VLAs such as $\pi_{0.5}$~\cite{black2025pi05} generalize across scenes and are adapted to whole-body action spaces; GR00T~\cite{bjorck2025gr00t} is an open foundation model for humanoids trained on real, synthetic and video data; $\Psi_0$~\cite{wei2026psi0} pre-trains on human egocentric video and then fine-tunes on a smaller amount of real humanoid data; and WholeBodyVLA~\cite{jiang2026wholebodyvla} learns a unified latent action space for loco-manipulation. These policies differ in their data, action spaces and controllers, but all of them condition on the current observation and a short proprioceptive history, so an event that has left the camera's view can no longer affect their actions. Memory has so far been added mostly to robots that do not walk, e.g.\ EgoMI~\cite{yu2025egomi} keeps past views for a wheeled semi-humanoid with an actuated head and SOMA~\cite{li2026soma} builds a spatial memory from a movable head camera; to our knowledge, no prior work studies memory for whole-body loco-manipulation, where the robot's own walking and turning move cues out of view. Concurrent work, POT-VLA~\cite{ren2026closing}, keeps a 3D record of where each task object is now but not what happened before; \method{} keeps the past views in which objects changed.

\subsection{Memory and Datasets for Robot Policies}
\label{sec:related_mem}
\emph{Memory mechanisms.} Memory for robot policies has been studied mainly on tabletop arms with fixed or wrist cameras~\cite{dai2026robomme,shi2026memoryvla,fang2025sam2act,sridhar2026memer}, and on mobile and navigation agents~\cite{cheng2025navila}. Long frame histories~\cite{jang2025contextvla,torne2026mem} grow in cost and dilute a short cue; compression into gist tokens~\cite{wang2026nativemem} or memory banks and object slots~\cite{shi2026memoryvla,chung2026rethinking} keeps the cost fixed but can lose the detail a decision needs; frame selection keeps native detail at a fixed budget, by uniform sampling~\cite{dai2026robomme}, retrieval~\cite{sridhar2026memer} or events~\cite{zeng2026kemo,yang2026eventvla}. A humanoid adds two constraints: its own turning and walking move a cue out of the head camera's view and change the viewpoint, and the memory must be updated causally while whole-body control runs. \method{} therefore selects frames by object events, measured relative to the scene after camera-motion removal, and reads them with action-query cross-attention and FiLM. Unlike SAM2Act+~\cite{fang2025sam2act}, which adopts SAM2's memory mechanism, we use SAM3 only as a tracker whose object events select the past views the action expert reads.
\emph{Datasets and benchmarks.} Memory benchmarks are mostly tabletop~\cite{dai2026robomme,lei2026robomemarena,fang2025sam2act,chung2026rethinking}, and humanoid benchmarks~\cite{sferrazza2024humanoidbench,wei2026simple,wang2026humanoidarena,lin2026humanoidmimicgen} evaluate whole-body skills with the goal in view. Our benchmark fills this gap with whole-body humanoid loco-manipulation tasks whose decisions depend on a cue that is no longer in view.

\section{Memory-Dependent Loco-Manipulation Benchmark}
\label{sec:benchmark}
\begin{table*}[t]
\vspace*{5pt}
\caption{Results on the 14 SIMPLE Tasks.}
\label{tab:simple14}
\centering\scriptsize
\setlength{\tabcolsep}{1.7pt}
\begin{tabular}{@{}lccccccccc@{}}
\toprule
Task & \begin{tabular}[b]{@{}c@{}}InternVLA-M1\\\cite{chen2025internvlam1}\end{tabular} & \begin{tabular}[b]{@{}c@{}}DP\\\cite{chi2023diffusion}\end{tabular} & \begin{tabular}[b]{@{}c@{}}ACT\\\cite{zhao2023learning}\end{tabular} & \begin{tabular}[b]{@{}c@{}}EgoVLA\\\cite{yang2025egovla}\end{tabular} & \begin{tabular}[b]{@{}c@{}}$\pi_{0.5}$\\\cite{black2025pi05}\end{tabular} & \begin{tabular}[b]{@{}c@{}}GR00T\\\cite{bjorck2025gr00t}\end{tabular} & \begin{tabular}[b]{@{}c@{}}$\Psi_0$\\\cite{wei2026psi0}\end{tabular} & \begin{tabular}[b]{@{}c@{}}Uniform\\memory\end{tabular} & \begin{tabular}[b]{@{}c@{}}\method{}\\(ours)\end{tabular} \\
\midrule
T1 XMovePick & 0 / 0 / 0 & 5 / 3 / 3 & 7 / 6 / 3 & 3 / 3 / 1 & 8 / 6 / 3 & 6 / 6 / 4 & \underline{9 / 10 / 6} & \underline{10 / 10 / 5} & \textbf{10 / 10 / 7} \\
T2 BendPickMP & 4 / 5 / 4 & \underline{10 / 9 / 10} & 9 / 9 / 9 & 6 / 9 / 6 & 8 / 8 / 6 & \underline{10 / 10 / 9} & \underline{10 / 10 / 9} & \textbf{10 / 10 / 10} & \textbf{10 / 10 / 10} \\
T3 Handover & 0 / 0 / 1 & 1 / 1 / 1 & 1 / 2 / 2 & 2 / 3 / 4 & 4 / 2 / 3 & 6 / 6 / 7 & 7 / 6 / 8 & \underline{8 / 7 / 9} & \textbf{10 / 9 / 9} \\
T4 TabletopGraspMP & 6 / 7 / 9 & 4 / 4 / 1 & \underline{10 / 10 / 8} & 6 / 5 / 4 & \textbf{10 / 10 / 9} & \underline{10 / 10 / 8} & 10 / 10 / 7 & 9 / 10 / 8 & \textbf{10 / 10 / 9} \\
T5 XMoveBendPick & 5 / 4 / 4 & 0 / 0 / 0 & 2 / 2 / 3 & 3 / 6 / 6 & 0 / 0 / 0 & 6 / 2 / 4 & 9 / 4 / 4 & \underline{8 / 6 / 10} & \textbf{9 / 9 / 8} \\
T6 LocomotionPickBetweenTables & 0 / 0 / 0 & 0 / 1 / 0 & 1 / 0 / 0 & 1 / 3 / 1 & 2 / 0 / 1 & 0 / 0 / 0 & 4 / 4 / 3 & \underline{4 / 5 / 4} & \textbf{8 / 6 / 7} \\
T7 PickAndPlaceAndHugContainer & 0 / 0 / 0 & 0 / 0 / 0 & 3 / 6 / 6 & 1 / 0 / 0 & 2 / 3 / 1 & 0 / 0 / 0 & 4 / 5 / 3 & \underline{6 / 4 / 6} & \textbf{7 / 6 / 5} \\
T8 OpenFaucet & 0 / 0 / 0 & \underline{2 / 3 / 6} & 0 / 0 / 1 & 0 / 0 / 0 & 4 / 2 / 0 & 0 / 0 / 0 & 4 / 2 / 4 & \underline{3 / 5 / 3} & \textbf{8 / 5 / 7} \\
T9 CloseDoor & 0 / 0 / 0 & \textbf{10 / 10 / 10} & \textbf{10 / 10 / 10} & 10 / 8 / 10 & \underline{10 / 10 / 9} & \textbf{10 / 10 / 10} & \textbf{10 / 10 / 10} & \textbf{10 / 10 / 10} & \textbf{10 / 10 / 10} \\
T10 OpenOven & 8 / 6 / 8 & 0 / 0 / 0 & 1 / 0 / 1 & 5 / 8 / 2 & 4 / 3 / 6 & 0 / 0 / 1 & 5 / 6 / 6 & \underline{7 / 8 / 9} & \textbf{9 / 8 / 10} \\
T11 OpenTrashCan & 0 / 0 / 0 & 4 / 7 / 7 & 4 / 4 / 7 & 3 / 5 / 4 & 7 / 6 / 5 & 2 / 3 / 4 & 4 / 7 / 7 & \underline{7 / 5 / 7} & \textbf{8 / 7 / 8} \\
T12 PushOfficeChair & 2 / 3 / 1 & 10 / 6 / 8 & \textbf{9 / 10 / 10} & \textbf{10 / 9 / 10} & 10 / 10 / 6 & 5 / 5 / 4 & \textbf{10 / 10 / 9} & \underline{8 / 9 / 10} & \textbf{9 / 10 / 10} \\
T13 BendHandover & 5 / 2 / 2 & 0 / 0 / 0 & 0 / 0 / 0 & 3 / 2 / 4 & 8 / 9 / 8 & 0 / 0 / 0 & \underline{10 / 8 / 9} & \underline{9 / 9 / 9} & \textbf{10 / 10 / 10} \\
T14 BendPick & 3 / 0 / 0 & 0 / 0 / 0 & 0 / 0 / 0 & 0 / 1 / 2 & 0 / 1 / 2 & 0 / 0 / 1 & 8 / 7 / 7 & \underline{9 / 10 / 7} & \textbf{9 / 10 / 8} \\
\midrule
$\Sigma$ (14 tasks) & \begin{tabular}[c]{@{}c@{}}89/420\\(21.2\%)\end{tabular} & \begin{tabular}[c]{@{}c@{}}136/420\\(32.4\%)\end{tabular} & \begin{tabular}[c]{@{}c@{}}176/420\\(41.9\%)\end{tabular} & \begin{tabular}[c]{@{}c@{}}169/420\\(40.2\%)\end{tabular} & \begin{tabular}[c]{@{}c@{}}206/420\\(49.0\%)\end{tabular} & \begin{tabular}[c]{@{}c@{}}159/420\\(37.9\%)\end{tabular} & \begin{tabular}[c]{@{}c@{}}295/420\\(70.2\%)\end{tabular} & \underline{\begin{tabular}[c]{@{}c@{}}323/420\\(76.9\%)\end{tabular}} & \textbf{\begin{tabular}[c]{@{}c@{}}365/420\\(86.9\%)\end{tabular}} \\
\bottomrule
\end{tabular}
\\[3pt]
\begin{minipage}{\textwidth}\scriptsize All columns are our evaluation of the 14 SIMPLE tasks with the same client. Cells: successes at randomization levels L0 / L1 / L2 (10 episodes each). Bold: best per task; underlined: second best. $\Psi_0$: our fine-tuned checkpoint; Uniform memory: \method{} without object tokens and with evenly spaced frames. On these tasks, \method{} uses the fixed seven-frame memory (without adaptive memory).\end{minipage}
\end{table*}

\subsection{Task Definition}
\label{sec:bench_setting}

We call a humanoid loco-manipulation task \emph{memory-dependent} when, in a dynamic scene, the observation available at the decision does not determine the correct action. Each episode has three phases: in the \emph{cue} phase, a scene change, such as a lamp that lights or a mug that is hidden, singles out one answer $y$ from a small set $\mathcal{Y}$ (two or three pads); in the \emph{retention} phase, the robot performs whole-body skills that are the same for every answer; and in the \emph{decision} phase, from a time $t_d$ on, a whole-body action must follow $y$. A cue can become unobservable in two ways: it can end, as when a light goes off or an object is hidden (\emph{scene-induced}), or it can leave the head camera's view as the robot turns and walks (\emph{motion-induced}). We require that (i) the change is no longer observable at $t_d$, whether scene-induced or motion-induced; (ii) neither the instruction nor the robot's motion before $t_d$ depends on $y$; and (iii) $y$ is balanced over $\mathcal{Y}$. A policy without memory then cannot infer $y$ from the current view (chance: $1/|\mathcal{Y}|$), so we score \emph{memory-decision accuracy} separately from manipulation success (Sec.~\ref{sec:bench_eval}).

The kind of change determines the memory a decision needs: \emph{what} changed (object memory, O), \emph{where} an object went (spatial, S), or \emph{when} and how often it changed (temporal, T), as in the reference, permanence and counting tasks of RoboMME~\cite{dai2026robomme}. Our six simulated tasks (Fig.~\ref{fig:tasks}) cover all three. A lamp flashes one to three times in FlashPad (T); a mug disappears behind one of three screens while a decoy mug leaves in HideMug (S); a ball is concealed under one of three cups, which are then swapped, in ShellGame (S); one of two coloured lamps lights up in LitLamp (O); and one of three pads blinks for 0.2\,s in BlinkPad and TouchPad (O). Afterwards, the robot puts the carried object on the indicated pad or, in TouchPad, touches that pad. Because the cues are short-lived or occluded~\cite{vanhoorick2023tracking,xia2026targo} while the robot watches, these suites isolate scene-induced loss; the real-world tasks additionally introduce motion-induced loss (Sec.~\ref{sec:exp_real}).

\subsection{Data Generation}
\label{sec:bench_gen}
A scripted whole-body expert generates the demonstrations in six steps.
\begin{enumerate}
  \item \emph{Task specification.} A script sets the single-table scene (the object to carry, the answer pads, the cue objects), the cue timeline and the skill sequence. A seed fixes the object pose ($\pm$3\,cm, $\pm$8\,cm, $\pm$0.15\,rad), the robot's spawn offset ($\pm$3\,cm), the cue layout and the appearance; the answer is drawn independently.
  \item \emph{Expert rollout.} In MuJoCo (2\,ms physics, 50\,Hz control), the cue plays for a fixed, task-specific window. In FlashPad, HideMug, ShellGame and LitLamp it starts 0.5\,s into the episode, while the robot still stands at the table before the grasp and lift; in BlinkPad it starts 0.5\,s after the lift, and in TouchPad at a random time 1.9--3.7\,s into the episode, before the robot turns. A precomputed Dex3 grasp, cuRobo arm planning and AMO~\cite{li2025amo} leg and waist balancing drive the robot through the skill sequence.
  \item \emph{Selection.} An episode is kept if the object stays held after the lift (in the five carrying tasks) and the cue is visible to the head camera during its window; kept episodes are balanced over the answers.
  \item \emph{Rendering.} Recorded states are replayed in Isaac Sim and rendered from the head camera ($640{\times}360$, 50\,Hz) with per-episode randomized background distractors, materials and lighting.
  \item \emph{Export.} Episodes are exported in the LeRobot v2.1 format (\mbox{32-D} state, \mbox{36-D} whole-body action, 50\,Hz video), each ending 1\,s after the placement.
  \item \emph{Training and test split.} Each memory task has 100 training demonstrations and 90 test scenes from seeds distinct from training. Test seeds pass the same selection as the training episodes and stay in distribution: layout and expert rollout within the training range, table materials seen in training. They are balanced over the answers and the three randomization levels.

\end{enumerate}

\begin{table}[t] 
\caption{Comparison with Robot Memory and Humanoid Benchmarks.}
\label{tab:data_stats}
\centering\scriptsize
\setlength{\tabcolsep}{1.6pt}
\begin{tabular}{@{}llcccccc@{}}
\toprule
Benchmark & Robot & \makecell{Whole-\\body} & Memory & Data & Tasks & Steps & \makecell{Mem.\\(\%)} \\
\midrule
MemoryBench~\cite{fang2025sam2act} & Franka & \ding{55} & \ding{51} & Sim & 3 & -- & -- \\
LIBERO-Mem~\cite{chung2026rethinking} & Franka & \ding{55} & \ding{51} & Sim & 10 & -- & -- \\
RoboMME~\cite{dai2026robomme} & Franka & \ding{55} & \ding{51} & Sim+Real & 16+4 & 481 & -- \\
RoboMemArena~\cite{lei2026robomemarena} & Dual-arm & \ding{55} & \ding{51} & Sim+Real & 26+5 & 1{,}076 & -- \\
\midrule
HumanoidMimicGen~\cite{lin2026humanoidmimicgen} & G1 & \ding{51} & \ding{55} & Sim & 9 & 818 & 0 \\
HumanoidArena~\cite{wang2026humanoidarena} & G1 & \ding{51} & \ding{55} & Sim & 7 & 611 & 0 \\
SIMPLE~\cite{wei2026simple} & G1 & \ding{51} & \ding{55} & Sim & 14 & 322 & 0 \\
\midrule
\textbf{HuMeGo (Ours)} & G1 & \ding{51} & \ding{51} & \textbf{Sim+Real} & \textbf{6+6} & \textbf{1{,}155} & \textbf{40.7} \\
\bottomrule
\end{tabular}
\\[3pt]
\begin{minipage}{\columnwidth}\scriptsize Whole-body: humanoid loco-manipulation. Memory: the decision depends on a vanished cue. Tasks: sim + real. Steps: mean episode length. Mem.: share of steps after the decision; --: not reported.\end{minipage}
\end{table}
\subsection{Data Analysis}
\label{sec:bench_analysis}

As shown in Table~\ref{tab:data_stats}, we contrast our benchmark with robot memory benchmarks and with humanoid imitation benchmarks collected on the same platform and control frequency (Unitree G1, 50\,Hz). Two properties set it apart. \emph{Extended horizons:} an average episode spans 1{,}155 control steps (23\,s), compared with 818, 611 and 322 steps in HumanoidMimicGen, HumanoidArena and SIMPLE, respectively. \emph{Decisions that require memory:} each task presents a short-lived cue that occupies only 0.8--14.7\% of an episode and has to be remembered for 13.5--22.5\,s after it disappears (task medians). Overall, 40.7\% of the steps come after the decision point, where, following conditions~(i) and~(ii) in Sec.~\ref{sec:bench_setting}, the vanished cue provides the sole evidence for the correct action. In contrast, no step of the humanoid benchmarks depends on a vanished cue, and the memory benchmarks use arms, not humanoids.
\begin{table*}[t]
\vspace*{5pt}
\caption{Memory Task Results.}
\label{tab:main}
\centering\footnotesize
\setlength{\tabcolsep}{4pt}
\begin{tabular}{@{}lccccccc@{}}
\toprule
 & Counting (T) & \multicolumn{2}{c}{Permanence (S)} & \multicolumn{3}{c}{Reference (O)} & \\
\cmidrule(lr){2-2}\cmidrule(lr){3-4}\cmidrule(lr){5-7}
Method & FlashPad & HideMug & ShellGame & LitLamp & BlinkPad & TouchPad & Avg. \\
\midrule
\multicolumn{8}{@{}l}{\textit{Success (L0 / L1 / L2)}} \\
$\Psi_0$ (no memory)~\cite{wei2026psi0} & 7 / 7 / 5 & 11 / 14 / 12 & 7 / 11 / 7 & 11 / 6 / 4 & \underline{11 / 11 / 11} & 3 / 11 / 8 & 8.3 / 10.0 / 7.8 \\
Uniform memory & \underline{22 / 23 / 22} & \underline{12 / 20 / 15} & \underline{10 / 9 / 9} & \underline{12 / 12 / 11} & 10 / 9 / 13 & \underline{6 / 13 / 12} & \underline{12.0 / 14.3 / 13.7} \\
\method{} (ours) & \textbf{26 / 27 / 23} & \textbf{21 / 20 / 20} & \textbf{21 / 20 / 19} & \textbf{18 / 15 / 12} & \textbf{26 / 22 / 25} & \textbf{26 / 25 / 26} & \textbf{23.0 / 21.5 / 20.8} \\
\midrule
\multicolumn{8}{@{}l}{\textit{Memory-decision accuracy (\%)}} \\
\textcolor{gray}{Chance} & 33 & 33 & 33 & 50 & 33 & 33 & 36 \\
$\Psi_0$ (no memory)~\cite{wei2026psi0} & 30 & 47 & 33 & 47 & \underline{45} & 32 & 39 \\
Uniform memory & \underline{82} & \underline{57} & \underline{35} & \underline{51} & 37 & \underline{38} & \underline{50} \\
\method{} (ours) & \textbf{97} & \textbf{99} & \textbf{84} & \textbf{100} & \textbf{87} & \textbf{95} & \textbf{94} \\
\bottomrule
\end{tabular}
\\[3pt]
\begin{minipage}{\textwidth}\footnotesize Successes at L0 / L1 / L2 (30 episodes per level); memory-decision accuracy as defined in Sec.~\ref{sec:bench_eval}.\end{minipage}
\end{table*}

\section{Method: \method{}}
\label{sec:method}
\subsection{Preliminaries: Whole-Body VLA Policies}
\label{sec:method_psi0}
A whole-body VLA policy $\pi_\theta(\mathbf{a}_t\mid I_t,s_t,\ell)$ maps the head-camera image $I_t$, proprioceptive state $s_t$ and instruction $\ell$ to $H$ future actions $\mathbf{a}_t=(a_t,\dots,a_{t+H-1})$~\cite{chi2023diffusion,chen2025covdiffusion} at 50\,Hz: upper-body joint targets ($q_{\mathrm{arm}}$, $q_{\mathrm{hand}}$) and a lower-body command (\{torso, $h_b$, $v$\}: torso orientation, base height and base velocity) that a lower-body controller turns into $q_{\mathrm{lower}}$ (Fig.~\ref{fig:pipeline}). Recent policies such as $\pi_{0.5}$~\cite{black2025pi05}, GR00T~\cite{bjorck2025gr00t} and $\Psi_0$~\cite{wei2026psi0} encode $I_t$ and $\ell$ with a vision-language backbone and generate $\mathbf{a}_t$ with a flow-matching action expert~\cite{lipman2023flow}, whose action tokens embed $s_t$ and the noise $\varepsilon$ (Fig.~\ref{fig:pipeline}). Conditioned only on $(I_t,s_t,\ell)$, they cannot respond to an event that has left the view. \method{} adds memory tokens $\mathbf{M}_t$ of the past views in which task objects changed and object tokens $\mathbf{O}_t$, giving $\pi_\theta(\mathbf{a}_t\mid I_t,s_t,\ell,\mathbf{M}_t,\mathbf{O}_t)$ (Secs.~\ref{sec:method_memory} and~\ref{sec:method_attention}); selecting views by object events keeps this memory small yet retains short-lived cues.

\subsection{Object-Guided Memory Selection}
\label{sec:method_memory}
The memory is built in three stages (Fig.~\ref{fig:pipeline}, left): SAM3 tracks the task objects, object events flag the frames where they change, and a causal selector fills an adaptive budget.

\begin{algorithm}[t]
\caption{Causal memory selection at time $t$}
\label{alg:selector}
\small
\begin{algorithmic}[1]
\Require event store $\mathcal{E}$ (frame $\mapsto$ scored events), tracks available at $t$, budget $B_{\max}$, gap $g{=}5$; $\mathrm{uniform}(I,k)$: $k$ evenly spaced frames of $I$ including both ends (the last frame if $k{=}1$)
\State add the events of the tracks available at $t$ to $\mathcal{E}$
\State $S\gets\mathrm{uniform}\big([0,t],\min(3,t{+}1)\big)$ \Comment{first, middle, current}
\State \textbf{if} no reliable track yet \textbf{or} no event frame $f<t$ \textbf{then return} $S$
\For{event frames $f<t$ by decreasing score, until $B_{\max}{-}3$ are added}
  \State \textbf{if} an object of $f$ has no frame in $S$ within $g$ \textbf{then} add $f$
\EndFor
\State \Return $S$ \Comment{3 to $B_{\max}$ frames}
\end{algorithmic}
\end{algorithm}

\textbf{Memory budget.} A memory frame keeps the $K{=}77$ native tokens ($7{\times}11$ grid) of the frozen vision tower, and the budget is $B_{\max}{=}7$ frames, the current and six past ones: 539 memory tokens plus at most 14 object tokens (Sec.~\ref{sec:method_attention}). Tokens are cached, so re-selecting a frame is free. Unlike a pooled summary, native tokens keep the small regions a decision may hinge on (a cue can cover under 1\% of the image), but only six past frames fit, so \emph{which} are kept matters.

\textbf{Object tracking.} The task decides what to remember: a frozen SAM3~\cite{carion2026sam} tracks the task objects over the episode so far ($o_{0:t}$ in Fig.~\ref{fig:pipeline}), prompted with one or two short phrases (e.g.\ ``blue cup'' in ShellGame, ``cracker box'' in Handover, T3 in Table~\ref{tab:simple14}), and returns a mask, a box and a confidence per frame and object. Prompts are written once per task by Claude Opus~5.5 and fixed before evaluation; the event rules and thresholds below are shared by all tasks. In simulation, SAM3 tracks every 50-Hz control step, and frame counts below refer to tracked frames. A similarity transform fitted to background features of consecutive frames removes camera motion, so object motion is measured relative to the scene.

\textbf{Object events.} A past frame becomes a memory candidate only if a tracked object in it appears, disappears (we keep its last visible frame), reappears, moves by over 3\% of the image diagonal after camera-motion removal, or changes shape (mask overlap with the previous one below 65\%). Events are ranked by type, disappearance and reappearance first, then by confidence; detections below 0.35 are ignored. For example, a blinking pad appears and a hidden mug disappears.

\begin{figure*}[t]
  \vspace*{5pt}
  \centering
  \includegraphics[width=\textwidth]{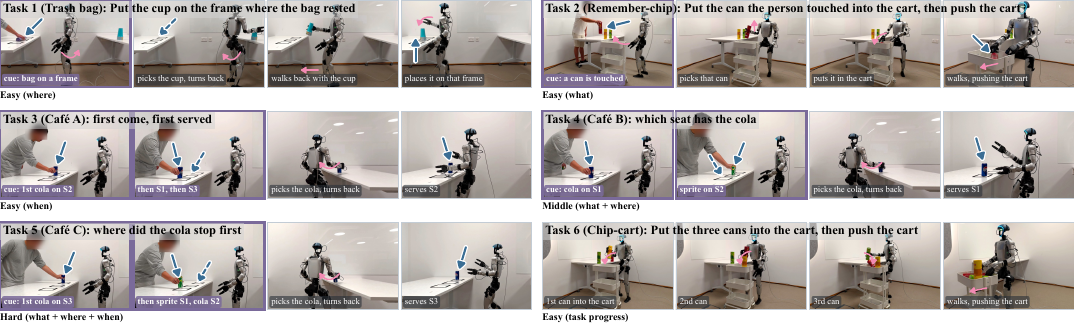}
  \caption{Real-world memory tasks on the Unitree G1. Each panel shows the task instruction and four steps of a teleoperated demonstration; the line below gives the difficulty and the memory the decision needs. Blue arrows mark the cue and the answer (dashed: where the cue was, after it is gone); pink arrows show the robot's motion in each step, straight for walking. S1--S3: the three seats of the caf\'e tasks, from front to back.}
  \label{fig:real}
\end{figure*}

\textbf{Causal selection.} At each time $t$ (Algorithm~\ref{alg:selector}), the first, middle and current frames are always kept, and up to four slots go to the top-scoring event frames, skipping one whose objects all have a selected frame within 5 frames. The remaining slots stay empty, so the memory holds 3 to 7 frames (5.4 on average); the variant without adaptive memory, used on SIMPLE, fills all seven. Only frames up to $t$ are eligible, so the memory never depends on later frames; seven evenly spaced frames form our Uniform memory control.

\subsection{Memory Cross-Attention in the Action Expert}
\label{sec:method_attention}
Memory enters the policy only through the action expert: its action tokens read $\mathbf{M}_t$ and $\mathbf{O}_t$ by cross-attention, and the read-out modulates them through FiLM~\cite{perez2018film}, as in the memory modulator of RoboMME~\cite{dai2026robomme}.

\textbf{Memory tokens.} As the selected frames are irregularly spaced, each native token $f_i\in\mathbb{R}^{2048}$ gets a sinusoidal code $\phi(\tau_i,p_i)$ of its frame index $\tau_i$ and grid cell $p_i$, and both are projected to the 1536-D width of the action expert:
\begin{equation}
m_i = W_m\,[\,f_i;\ \mathrm{SiLU}(W_p\,\phi(\tau_i,p_i))\,].
\label{eq:memtok}
\end{equation}
Empty slots are masked; $\mathbf{M}_t=[m_1,\dots,m_{539}]$ is computed once per policy call and shared by all blocks and flow-matching steps.

\textbf{Memory retrieval.} In each of the six blocks $j$, after joint attention, the $H{=}30$ action tokens query the memory with their RMS-normalized form $\bar{\mathbf{A}}_j$:
\begin{equation}
\mathbf{c}_j=\mathrm{Attn}\big(\bar{\mathbf{A}}_j W_Q,\ \mathbf{M}_tW_K,\ \mathbf{M}_tW_V\big)\,W_O,
\label{eq:memattn}
\end{equation}
(multi-query attention; rotary codes place the action steps after the packed memory tokens), so each future action step retrieves the frames it needs while the pretrained joint attention stays unchanged.

\textbf{Modulation.} Rather than entering the residual stream, the read-out scales and shifts the normalized action tokens ahead of the block's adaptive layer norm and feed-forward network:
\begin{equation}
[\boldsymbol{\gamma}_j,\boldsymbol{\beta}_j]=W_F\,\mathbf{c}_j,\quad
\tilde{\mathbf{A}}_j=\bar{\mathbf{A}}_j\odot(1+\boldsymbol{\gamma}_j)+\boldsymbol{\beta}_j,
\label{eq:film}
\end{equation}
With $W_F$ initialized near zero, the policy initially matches the pretrained one.

\textbf{Object tokens.} The SAM3 masks also give an object-level view (Object Masking in Fig.~\ref{fig:pipeline}). For up to two tracked objects per memory frame, the native tokens are averaged with each cell's mask coverage as weight and combined with a sinusoidal code of frame index and mask centroid, the box, area and visibility, and a learned slot embedding, giving up to 14 object tokens $\mathbf{O}_t$. A second cross-attention adds their read-out before Eq.~\eqref{eq:film}, $\mathbf{c}_j \leftarrow \mathbf{c}_j + g\,\mathrm{Attn}(\bar{\mathbf{A}}_j,\mathbf{O}_t)$, with a learnable $g$ (initially 1) and a zero-initialized output projection, so the branch starts silent yet trainable.

\textbf{Decision head.} To map the remembered cue to a pad, a learned query attends over the memory and object tokens and a null token, and an MLP outputs one logit per answer, trained with a cross-entropy loss on the demonstrated answer from the decision step on. The softmax of these logits, the dynamic weights $w$ (Fig.~\ref{fig:pipeline}), enters every block's adaptive layer norm through a zero-initialized map on the time embedding, so training starts from the head-free policy.

\textbf{Pad tokens.} The weights $w$ also point the action expert at a pad. Each pad visible in the current frame gives one token (its native tokens pooled over the SAM3 box, the box geometry and a pad index), forming $\mathbf{P}_t$. In every block, a further cross-attention with a zero-initialized output reads $w\odot\mathbf{P}_t$ by adding $\log w_k$ to the score of pad $k$ and adds its read-out to $\mathbf{c}_j$, so a confident $w$ hands the chosen pad's token to the action tokens; the branch is active only when the largest weight and the weight on visible pads reach 0.6.

\subsection{Training Details}
\label{sec:method_training}
Each policy is fine-tuned per task from the pretrained base policy (Sec.~\ref{sec:exp_setup}) with $\mathcal{L}=\mathbb{E}\,\lVert v_\theta(\mathbf{a}_t^\sigma,\sigma)-(\varepsilon-\mathbf{a}_t)\rVert^2+\lambda\,\mathcal{L}_{\mathrm{CE}}$: the flow-matching loss of SIMPLE~\cite{wei2026simple} and $\Psi_0$~\cite{wei2026psi0}, where the action expert $v_\theta$ sees $\mathbf{a}_t^\sigma=(1-\sigma)\,\mathbf{a}_t+\sigma\varepsilon$ with $\sigma\sim\mathcal{U}(0,1)$, plus the decision head's cross-entropy ($\lambda{=}0.5$). The Qwen3-VL-2B backbone~\cite{bai2025qwen3vl} stays frozen; the memory encoder, the memory attention (6 query heads of 256 dimensions, one shared key/value head) and $W_F$ (standard deviation 0.002 at initialization) are trained from scratch. We use AdamW ($\beta{=}(0.95,0.999)$), a learning rate of $10^{-4}$ with cosine decay and 1k warm-up steps, bf16 mixed precision, dropout 0.1 and a global batch of 128, for 40k steps on the 94--100 demonstrations of a SIMPLE task or the 100 of a memory task. On the real robot, whose action space differs, each task is trained separately on its 15 teleoperated demonstrations: the action expert's transformer blocks start from the post-trained action expert, the memory modules are trained from scratch, and the 5k-step checkpoint is deployed.

\section{Experiments}
\label{sec:exp}
Our experiments answer four questions:
\begin{enumerate}
  \item[\textbf{Q1}] Does adding memory cost reactive skills on the 14 standard SIMPLE tasks, or even improve them?
  \item[\textbf{Q2}] On our six memory tasks, does \method{} act on vanished cues that memoryless policies miss?
    \item[\textbf{Q3}] What is the role of pre-history memory: does object guidance, not capacity, explain the gain, and what do key frames, object tokens, the decision head, adaptive memory and memory attention each contribute?
  \item[\textbf{Q4}] Does memory carry over to a real Unitree G1, where the robot also turns away from the cue?
\end{enumerate}

\subsection{Setup}
\label{sec:exp_setup}
\textbf{Simulation.} All simulated experiments use SIMPLE~\cite{wei2026simple} with the robot, \mbox{36-D} whole-body action and rendering of Sec.~\ref{sec:bench_gen}. AMO~\cite{li2025amo} drives the lower body in the memory tasks and in the two SIMPLE motion-planning tasks (T2, T4), and SIMPLE's decoupled whole-body controller drives it in the other 12 SIMPLE tasks, whose demonstrations are teleoperated. Each policy runs as a server that the unmodified SIMPLE evaluation client, which also runs our memory tasks, queries once per action chunk.

\textbf{Real robot.} We use a Unitree G1 (29 DoF) with Dex3 hands and a head-mounted RealSense D435 ($640{\times}360$, 30\,Hz). The policy runs on one RTX 5090 with real-time chunking~\cite{black2025realtime}: of the $H{=}30$ predicted actions, the first 24 are executed before the next chunk takes over; SONIC~\cite{luo2025sonic} executes its \mbox{64-D} motion token and \mbox{14-D} hand targets at 50\,Hz, and SAM3 tracks one frame per executed chunk (every 24 frames, about 1.25\,Hz).

\textbf{Baselines.} All baselines are trained by us on the same demonstrations. (i) \emph{Policies without memory}: InternVLA-M1~\cite{chen2025internvlam1}, Diffusion Policy~\cite{chi2023diffusion}, ACT~\cite{zhao2023learning}, EgoVLA~\cite{yang2025egovla}, $\pi_{0.5}$~\cite{black2025pi05}, GR00T~\cite{bjorck2025gr00t} and $\Psi_0$~\cite{wei2026psi0}, the memoryless base policy. (ii) \emph{Uniform memory} (RoboMME-style): as no published memory method exists for whole-body humanoid VLAs, we adapt the best-performing memory of RoboMME~\cite{dai2026robomme}, evenly sampled frames read through a memory modulator (FrameSamp + Modul), to our tasks. It is the Uniform memory control of Sec.~\ref{sec:method_memory}, i.e.\ \method{} with evenly spaced frames and no object tokens or decision head, at the same 539-token budget (512 in RoboMME), with native instead of pooled tokens, as a cue can cover under 1\% of the image; in an earlier evaluation on T5, T6 and T8, RoboMME's original 32 pooled frames of 16 tokens scored the same as this adaptation (both 53.3\%). On the real robot, we compare with $\Psi_0$ (Sec.~\ref{sec:exp_real}).

\textbf{Evaluation.}\label{sec:bench_eval} Every policy follows the standardized protocol of SIMPLE~\cite{wei2026simple} at three randomization levels (L0: visual and distractor randomization; L1: plus lighting; L2: plus spatial perturbation). The 14 SIMPLE tasks, which need no cue memory, use 10 episodes per level; our six memory tasks use 30 episodes per level (balanced answers; Sec.~\ref{sec:bench_gen}). An episode succeeds if its final condition holds (e.g.\ the object lies on the correct pad). On the memory tasks we also report \emph{memory-decision accuracy}, the fraction of episodes in which the decision head picks the cued answer, otherwise the fraction of placements or touches on the cued pad.

\begin{table}[!h]
\caption{Ablations on the Memory Tasks.}
\label{tab:ablation}
\centering\footnotesize
\setlength{\tabcolsep}{4pt}
\begin{tabular}{@{}lcc@{}}
\toprule
 & Successes & Success \\
Method & L0 / L1 / L2 & (\%) \\
\midrule
Uniform memory & 12.0 / 14.3 / 13.7 & 44.4 \\
\method{} w/o object tokens \& head & 18.0 / 19.0 / 17.3 & 60.4 \\
\method{} w/o decision head & 20.7 / 18.7 / 19.8 & 65.7 \\
\method{} with memory as context & 8.3 / 7.0 / 8.0 & 25.9 \\
\method{} w/o adaptive memory & 21.7 / 21.3 / 20.3 & 70.4 \\
\method{} (ours) & \textbf{23.0 / 21.5 / 20.8} & \textbf{72.6} \\
\bottomrule
\end{tabular}
\\[3pt]
\begin{minipage}{\columnwidth}\footnotesize Averages over the six tasks: successes per level (of 30 episodes). Uniform memory as in Table~\ref{tab:main}; the ablations remove adaptive memory, the object tokens together with the decision head, the decision head alone, or the memory attention, replaced by memory tokens in the VLM input (memory as context); the last three are applied to the variant without adaptive memory.\end{minipage}
\end{table}

\subsection{Results on Standard Loco-Manipulation Tasks (Q1)}
\label{sec:exp_main}
\textbf{Memory does not cost reactive skills.} On the 14 SIMPLE tasks (Table~\ref{tab:simple14}), \method{} matches or exceeds the best baseline on every task and reaches 86.9\% success, against 70.2\% for $\Psi_0$ and 76.9\% for Uniform memory. These tasks need no cue memory, but the moving head camera loses targets as the robot walks and bends, and one frame hides task progress; past frames fill these gaps, and object tokens locate objects more finely. Accordingly, the gain over Uniform memory is largest where the robot walks to or between its targets: T6 (21 vs.\ 13) and T8 (20 vs.\ 11) give 17 of the 42 extra successes, while both solve T2 and T9.
\subsection{Results on Memory-Dependent Tasks (Q2)}
\label{sec:exp_memory}
\textbf{\method{} acts on the cue; memoryless policies guess.} On the six memory tasks (Table~\ref{tab:main}), $\Psi_0$ decides near chance and Uniform memory, the best-performing memory of RoboMME~\cite{dai2026robomme} adapted to our tasks, is clearly above chance only on FlashPad and HideMug, as evenly spaced frames miss brief cues, whereas \method{} identifies the cue in 93.7\% of its decisions. Success rises to 72.6\%, against 29.1\% ($\Psi_0$) and 44.4\% (Uniform memory), and stays stable across randomization levels.

\subsection{Object-Guided vs.\ Uniform Memory (Q3)}
\label{sec:exp_ablation}
\textbf{Object guidance, not capacity, explains the gain.} At Uniform memory's budget (seven frames, 539 tokens), choosing frames by object events alone (\method{} w/o object tokens, which lacks the decision head) lifts success from 44.4\% to 60.4\% (Table~\ref{tab:ablation}), and adaptive memory loses nothing with 22.9\% fewer tokens: the role of pre-history memory is to keep the right frames, not more.

\textbf{Every part contributes.} Of the 28.2-point gain over Uniform memory, key frames give 16.0, object tokens 5.3, the decision head 4.7 and adaptive memory 2.2 (Table~\ref{tab:ablation}): key frames keep brief cues that evenly spaced frames miss, object tokens identify which object changed, and the decision head maps the cue to a target pad, so its answer labels explain only about a sixth of the gain. Memory fed to the frozen VLM as context instead of through the memory attention falls below even the memoryless $\Psi_0$.

\begin{table}[!h]
\caption{Real-World Results on the Unitree G1.}
\label{tab:real}
\centering\footnotesize
\setlength{\tabcolsep}{6pt}
\begin{tabular}{@{}llcc@{}}
\toprule
Task & Mem. & $\Psi_0$~\cite{wei2026psi0} & \method{} \\
\midrule
Trash bag & S & 4/10 & \textbf{9/10} \\
Remember-chip & O & 3/10 & \textbf{10/10} \\
Caf\'e A & T & 4/10 & \textbf{9/10} \\
Caf\'e B & O+S & 3/10 & \textbf{9/10} \\
Caf\'e C & O+S+T & 3/10 & \textbf{8/10} \\
Chip-cart & P & 5/10 & \textbf{8/10} \\
\midrule
Total & & 22/60 & \textbf{53/60} \\
\bottomrule
\end{tabular}
\\[3pt]
\begin{minipage}{\columnwidth}\footnotesize Successes over 10 trials per task and policy with memory types: object (O), spatial (S), temporal (T) memory or task progress (P).\end{minipage}
\end{table}

\subsection{Real-World Experiments (Q4)}
\label{sec:exp_real}
We deploy \method{} and $\Psi_0$ on the G1, with 10 trials per task and policy and balanced answers (Table~\ref{tab:real}).

\textbf{Tasks.} Fig.~\ref{fig:real} shows six tasks, graded by the memory types (what, where, when) a decision combines. In Tasks~1 and 3--5, a person gives the cue at one of three marked areas (``frames'') or seats on table~B; the robot then turns to table~A for the object and back, with the same motion for every answer, so the cue is out of view when it acts. In Task~2, the person touches one of the cans, leaving no trace, and in Chip-cart (Task~6) the robot must remember its progress once the remaining cans leave the view.

\textbf{Data.} For each task, we record 15 demonstrations (90 in total) by whole-body teleoperation (PICO headset, controllers and ankle trackers driving SONIC~\cite{luo2025sonic} at 50\,Hz; Sec.~\ref{sec:method_training}), with every answer demonstrated five times in the cue tasks; each frame pairs the head-camera image and robot state with the SONIC motion token and hand targets as the action label.

\textbf{Memory carries over to the real robot.} Over all six tasks (Table~\ref{tab:real}), \method{} succeeds in 88.3\% of the trials (53/60), against 36.7\% (22/60) for $\Psi_0$, even when a decision combines what, where and when.

\section{Limitations and Conclusion}
\label{sec:conclusion}
We studied the role of pre-history memory, the object events seen before the robot acts on its goal, in humanoid loco-manipulation, with \method{}, an object-guided perceptual memory for whole-body VLAs, and a humanoid memory benchmark. Object events select the few past views that matter, which the action expert reads within an adaptive budget: keeping the right frames matters more than keeping more, so the policy acts on vanished cues in simulation and on a real G1 without losing reactive skills. SAM3 misses a cue whose object is never detected, online tracking adds inference cost, rendering leaves a sim-to-real gap, and the simulated demonstrations are scripted. Remembering before acting on the goal is a first step toward humanoids that act on what happened, not only on what is visible.

\end{document}